\documentclass[sigconf]{acmart}

\AtBeginDocument{%
  \providecommand\BibTeX{{%
    \normalfont B\kern-0.5em{\scshape i\kern-0.25em b}\kern-0.8em\TeX}}}

\setcopyright{acmlicensed}

\copyrightyear{2024}
\acmYear{2024}
\setcopyright{rightsretained}
\acmConference[KDD '24]{Proceedings of the 30th ACM SIGKDD Conference on Knowledge Discovery and Data Mining}{August 25--29, 2024}{Barcelona, Spain}
\acmBooktitle{Proceedings of the 30th ACM SIGKDD Conference on Knowledge Discovery and Data Mining (KDD '24), August 25--29, 2024, Barcelona, Spain}\acmDOI{10.1145/3637528.3672064}
\acmISBN{979-8-4007-0490-1/24/08}

\usepackage{amsmath}
\usepackage{graphicx}
\usepackage{multirow}
\usepackage{stfloats}
\usepackage{enumitem}
\usepackage{float}

\begin{document}

\title{MacroHFT: Memory Augmented Context-aware Reinforcement Learning  On High Frequency Trading}

\author{Chuqiao Zong}
\affiliation{%
  \institution{Nanyang Technological University}
  \country{Singapore}
}
\email{ZONG0005@e.ntu.edu.sg}

\author{Chaojie Wang}
\authornote{Corresponding Author}
\affiliation{%
  \institution{Skywork AI}
  \country{Singapore}}
\email{chaojie.wang@kunlun-inc.com}

\author{Molei Qin}
\affiliation{%
  \institution{Nanyang Technological University}
  \country{Singapore}
}
\email{molei001@e.ntu.edu.sg}

\author{Lei Feng}
\affiliation{%
 \institution{Singapore University of Technology and Design}
 \country{Singapore}}
 \email{feng_lei@sutd.edu.sg}

\author{Xinrun Wang}
\affiliation{%
  \institution{Nanyang Technological University}
  \country{Singapore}
}
\email{xinrun.wang@ntu.edu.sg}

\author{Bo An}
\affiliation{%
  \institution{Nanyang Technological University}
  \institution{Skywork AI}
  \country{Singapore}
}
\email{boan@ntu.edu.sg}

\renewcommand{\shortauthors}{Zong, et al.}
\def\bb{\textcolor{blue}}

\begin{abstract}
  High-frequency trading (HFT) that executes algorithmic trading in short time scales, has recently occupied the majority of cryptocurrency market. Besides traditional quantitative trading methods, reinforcement learning (RL) has become another appealing approach for HFT due to its terrific ability of handling high-dimensional financial data and solving sophisticated sequential decision-making problems, \emph{e.g.,} hierarchical reinforcement learning (HRL) has shown its promising performance on second-level HFT by training a router to select only one sub-agent from the agent pool to execute the current transaction.
  However, existing RL methods for HFT still have some defects: 
  1) standard RL-based trading agents suffer from the overfitting issue, preventing them from making effective policy adjustments based on financial context; 
  2) due to the rapid changes in market conditions, investment decisions made by an individual agent are usually one-sided and highly biased, which might lead to significant loss in extreme markets.
  To tackle these problems, we propose a novel Memory Augmented Context-aware Reinforcement learning method On HFT, \emph{a.k.a.} MacroHFT, which consists of two training phases: 
  1) we first train multiple types of sub-agents with the market data decomposed according to various financial indicators, specifically market trend and volatility, where each agent owns a conditional adapter to adjust its trading policy according to market conditions;
  2) then we train a hyper-agent to mix the decisions from these sub-agents and output a consistently profitable meta-policy to handle rapid market fluctuations, equipped with a memory mechanism to enhance the capability of decision-making.
  Extensive experiments on various cryptocurrency markets demonstrate that MacroHFT can achieve state-of-the-art performance on minute-level trading tasks. Code has been released in \url{https://github.com/ZONG0004/MacroHFT}.
  
  
  



\end{abstract}




\maketitle

\section{Introduction}

The financial market, which involves over 90 trillion dollars of market capacity, has attracted a massive number of investors. Among all possible assets, the cryptocurrency market has gained particular favor in recent years due to its high volatility, offering opportunities for rapid and substantial profit, and its around-the-clock trading capacity, which allows for greater flexibility and the opportunity for traders to react immediately \cite{chuen2017cryptocurrency, fang2022cryptocurrency}. 
To fully exploit the profit potential, high-frequency trading (HFT), a form of algorithmic trading executed at high speeds, has occupied the majority of cryptocurrency markets \cite{almeida2023systematic}.
Besides rule-based trading strategies designed by experienced human traders, reinforcement learning (RL) has emerged as another promising approach recently due to its terrific ability to handle high-dimensional financial data and solve complex sequential decision-making problems \cite{deng2016deep, zhang2020deep, liu2020finrl}.
However, although RL has achieved great performance in low-frequency trading \cite{deng2016deep,zhu2022quantitative,theate2021application}, there remains a technical gap in developing effective high-frequency trading algorithms for cryptocurrency markets because of long trading horizons and volatile market fluctuations.


Specifically, existing RL-based HFT algorithms for cryptocurrency trading still suffer from some drawbacks, mainly including: 1) most of the current methods tend to treat the cryptocurrency market as a uniform and stationary entity \cite{briola2021deep, jia2019quantitative} or distinguish market conditions only based on market trends \cite{qin2023earnhft}, neglecting the market volatility. 
This oversight is significant in highly dynamic cryptocurrency markets. Ignoring the differences between markets with varying volatility levels can result in poor risk management and reduce the proficiency and specialization of trading strategies;
2) previous work \cite{zhang2023towards} indicates that 
existing strategies often suffer from overfitting, focusing on a small fraction of market features and disregarding recent market conditions, limiting their ability to adjust policies effectively based on the financial context;
3) individual agents' trading policies may fail to adjust promptly during sudden fluctuations, especially with large time granularity (e.g., minute-level trading tasks), which are common in cryptocurrency markets.

To tackle these aforementioned challenges, we propose a novel Memory Augmented Context-aware Reinforcement Learning on HFT, termed MacroHFT, focusing on minute-level cryptocurrency trading and incorporating macro market information as context to assist trading decision-making.
Specifically, the workflow of MacroHFT mainly consists of two phases: 1) in the first phase, 
MacroHFT decomposes the cryptocurrency market into different categories based on trend and volatility indicators. Multiple diversified sub-agents are then trained on different market dynamics, each featuring a conditional adapter to adjust its trading policy according to market conditions;
2) in the second phase, MacroHFT trains a hyper-agent as a policy mixture of all sub-agents, leveraging their profiting abilities under various market dynamics. The hyper-agent is equipped with a memory mechanism 
to learn from recent experiences, generating a stable trading strategy while maintaining the ability to respond to extreme fluctuations rapidly.




The main contributions of this paper can be summarized as:
\begin{enumerate}[leftmargin=1cm]
    \item We introduce a market decomposition method using trend and volatility indicators to enhance the specialization of sub-agents trained on decomposed market data. 
    
    \item We propose low-level policy optimization with conditional adaptation 
    for sub-agents, enabling efficient adjustments of trading policies according to market conditions. 
    
    \item We develop a hyper-agent that provides a meta-policy to effectively integrate diverse low-level policies from sub-agents. Utilizing a memory module, the hyper-agent can formulate a robust trading strategy by learning from highly relevant experiences.

    \item Comprehensive experiments on 4 popular cryptocurrency markets demonstrate that MacroHFT can significantly outperform many existing state-of-the-art baseline methods in minute-level HFT of cryptocurrencies.
\end{enumerate}

\section{Related Works}

In this section, we will give a brief introduction to the existing quantitative trading methods based on either traditional financial technical analysis or RL-based agents.



\subsection{Traditional Financial Methods}
Based on the assumption that past price and volume information can reflect future market conditions, technical analysis has been widely applied in traditional finance trading \cite{murphy1999technical}, and quantitative traders have designed millions of technical indicators as signals to guide the trading execution \cite{kakushadze2016101}.
For instance, Imbalance Volume (IV) \cite{chordia2002order} is developed to measure the difference between the number of buy orders and sell orders, which provides a clue of short-term market direction. Moving Average Convergence Divergence (MACD) \cite{hung2016various,krug2022enforcing} is another widely used trend-following momentum indicator showing the relationship between two moving averages of an asset's price, which reflects the future market trend. 

However, these traditional finance methods solely based on technical indicators often produce false trading signals in non-stationary markets like cryptocurrency, which may lead to poor performance, which has been criticized in recent studies \cite{liu2020adaptive, qin2023earnhft, li2019deep}.

\subsection{RL-based Methods}
Other than traditional finance methods, 
reinforcement learning based trading approaches have recently been another appealing approach in the field of quantitative trading. Besides directly applying standard deep RL algorithms like Deep-Q Network (DQN) \cite{mnih2015human} and Proximal Policy Optimization (PPO) \cite{schulman2017proximal}, various techniques were used as enhancements. CDQNRP \cite{zhu2022quantitative} generates trading strategies by applying a random perturbation to increase the stability of DQN training. CLSTM-PPO \cite{zou2024novel} applies LSTM to enhance the state representation of PPO for high-frequency stock trading. DeepScalper \cite{sun2022deepscalper} uses a hindsight bonus reward and auxiliary task to improve the agent's foresight and risk management ability.

Furthermore, to improve the adaptation capacity over long trading horizons containing different market dynamics, Hierarchical Reinforcement Learning (HRL) structures have also been applied to quantitative trading. 
HRPM \cite{wang2021commission} formulates a hierarchical framework to handle portfolio management and order execution simultaneously. MetaTrader \cite{niu2022metatrader} trains multiple policies using different expert strategies and selects the most suitable one based on the current market situation for portfolio management. EarnHFT \cite{qin2023earnhft} trains low-level agents under different market trends with optimal action supervisors and a router for agent selection to achieve stable performance in high-frequency cryptocurrency trading. 

However, the performance of existing HRL methods suffers from varying degrees of overfitting problems and has difficulty in making effective policy adjustments based on financial context, where MetaTrader \cite{niu2022metatrader} and EarnHFT \cite{qin2023earnhft}  only choose an individual agent to perform trading at each timestamp, usually leading to one-sided and highly biased decision execution.
To solve these challenges, we develop MacroHFT, which is the first HRL framework that not only incorporates macro market information as context to assist trading decision-making, but also provides a mixed policy to leverage sub-agents' specialization capacity by decomposing markets using multiple criteria, rather than selecting an individual one.



{
\setlength{\abovecaptionskip}{2pt}
\setlength{\belowcaptionskip}{0pt}
\begin{figure}[t]
\centering
\includegraphics[width=0.45\textwidth]{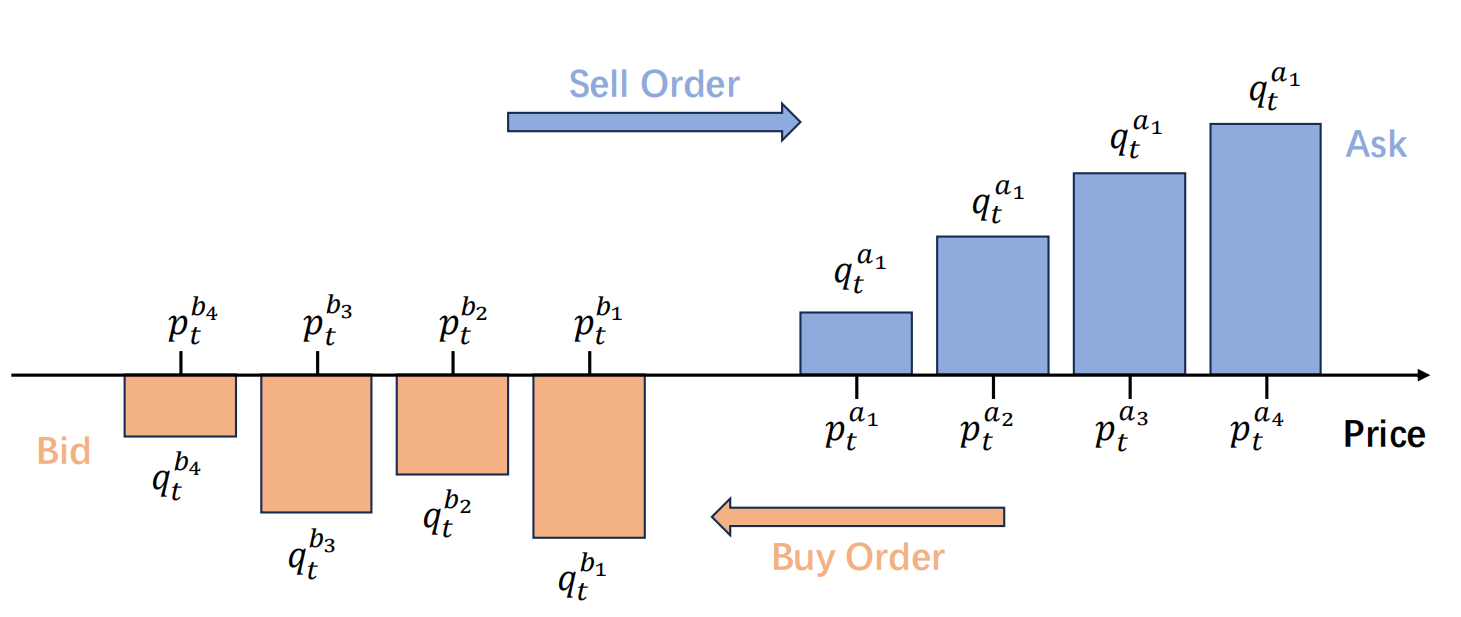}
\centering
\caption{A Snapshot of Limit Order Book (LOB)}
\label{fig:lob}
\end{figure}
}

\section{PRELIMINARIES}
In this section, we will first present the basic financial definitions that are related to cryptocurrency trading, and then elaborate our framework of hierarchical Markov Decision Process (MDP) structure that is different from previous works and focused on tackling minute-level high-frequency trading (HFT).

\subsection{Financial Definitions}
\label{Sec:3.1}

The common financial definitions of terms in HFT have been elaborated as follows:

\noindent \textbf{Limit Order} is an order placed by a market participant who wants to buy (bid) or sell (ask) a specific quantity of cryptocurrency at a specified price, where $(p^b, q^b)$ denotes a limit order to buy a total amount of $q^b$ cryptocurrency at the price $p^b$, and $(p^a, q^a)$ denotes a limit order of selling.

\noindent \textbf{Limit Order Book} (LOB), as shown as Fig~\ref{fig:lob}, serves as an important snapshot to describe the micro-structure of current market~\cite{madhavan2000market}, which is the record that aggregates buy and sell limit orders of all market participants for a cryptocurrency at the same timestamp \cite{rocsu2009dynamic}. 
Formally, we denote an $M$-level LOB ($M=5$ in our dataset) at time $t$ as $b_t=\{(p_t^{b_i}, q_t^{b_i}), (p_t^{a_i}, q_t^{a_i})\}_{i=1}^{M}$, where \(p_t^{b_i}, p_t^{a_i}\) denote the $i$-th level of bid and ask prices respectively, and \(q_t^{b_i}, q_t^{a_i}\) are the corresponding quantity for trading.


\noindent \textbf{Open-High-Low-Close-Volume} (OHLCV) is the aggregated information of executed market orders. At the timestamp $t$, the OHLCV information can be denoted as \(x_t=(p_t^{o}, p_t^{h}, p_t^{l}, p_t^{c}, v_t)\), where \(p_t^{o}, p_t^{h}, p_t^{l}, \allowbreak p_t^{c}\) denote the open, high, low, close prices and $v_t$ is the corresponding total volume of these market orders.

\noindent \textbf{Technical Indicators} are a group of features calculated from original LOB and OHLCV information by formulaic combinations, which can uncover the underlying patterns of the financial market. We denote the set of technical indicators at time $t$ as \(y_t=\phi(x_t, b_t, ..., \allowbreak x_{t-h+1}, b_{t-h+1})\), where \(h\) is the backward window length and \(\phi\) is the indicator calculator. Detailed calculation formulas of technical indicators used in our MacroHFT are provided in Appendix~\ref{appendix}.

\noindent \textbf{Position} is the amount of cryptocurrency a trader holds at a certain time $t$, which is denoted as \(P_t\), where \(P_t \geq 0\), indicating that only long position is allowed in our trading approach. 

\noindent \textbf{Net Value} is the sum of cash and the market value of cryptocurrency held by a trader, which can be calculated as \(V_t=V_{ct}+P_t \times p_t^{c}\), where \(V_{ct}\) is the cash value and \(p_t^{c}\) is the close price at timestamp $t$.

We highlight that the purpose of high-frequency trading is to maximize the final net value $V_t$ after executing market orders on a single cryptocurrency over a continuous period of time.

\subsection{MDP Formulation}
\label{Sec:3.2}
Due to the fact that high-frequency trading for cryptocurrency can be treated as a sequential decision-making problem, we can formulate it as an MDP constructed by a tuple \(<S, A, T, R, \gamma>\). To be specific, \(S\) is a finite set of states and \(A\) is a finite set of actions; \(T: S \times A \times S \rightarrow [0,1]\) is a state transition function which is composed of a set of conditional transition probabilities between states based on the taken action; \(R: S \times A \rightarrow \mathbb{R}\) is a reward function measuring the immediate reward of taking an action in a state; \(\gamma \in [0,1)\) is the discount factor. 
Then, a policy \(\pi: S \times A \rightarrow [0,1]\) will assign each state \(s \in S\) a distribution over action space \(A\), where \(a \in A\) has probability \(\pi(a|s)\). The objective is to find the optimal policy \(\pi^*\) so that the expected discounted reward $J = E_\pi \left[\sum_{t=0}^{+\infty} \gamma^t R_t\right]$ can be maximized.

When applying RL-based trading strategy for HFT, a single agent usually fails to learn an effective policy that can be profitable over a long time horizon because of the non-stationary characteristic in cryptocurrency markets.
To solve this problem, 
previous work \cite{qin2023earnhft} has shown that formulating HFT as a hierarchical MDP could be an effective solution on second-level HFT, where the low-level MDP operating on second-level time scale formulates trading execution under different market trends and the high-level MDP formulates strategy selection. 
Moving beyond second-level HFT, in this work, we focus on constructing a hierarchical MDP for minute-level HFT, where the low-level MDP formulates the process of executing actual trading under different types of market dynamics segmented by multiple criteria and the high-level MDP formulates the process of aggregating different policies through incorporating macro market information to construct a meta-trading strategy. 

Specifically, in our work, the hierarchical MDPs are operated under the same time scale (minute-level) so that the meta-policy can adapt more flexibly to frequent market fluctuations,  
which can be formulated as \((MDP_l, MDP_h)\)
\begin{align*}
    \begin{split}
        & MDP_l= <S_l, A_l, T_l, R_l, \gamma_l>, \\
        & MDP_h= <S_h, A_h, T_h, R_h, \gamma_h>
    \end{split}
\end{align*}

\noindent \textbf{Low-level State}, denoted as \(s_{lt} \in S_l \) at time $t$, consists of three parts: single state features $s_{lt}^1$, low-level context features $s_{lt}^2$ and position state $P_t$, where \(s_{lt}^1=\phi_1(x_t, b_t)\) denotes single-state features calculated from LOB and OHLCV snapshot of the current time step, \(s_{lt}^2=\phi_2(x_t, b_t, ..., x_{t-h+1}, b_{t-h+1})\) denotes context features calculated from all LOB and OHLCV information in a backward window of length $h=60$, \(P_t\) denotes the current position of the agent.

\noindent \textbf{Low-level Action} \(a_{lt} \in \{0,1\}\) is the action of sub-agent which indicates the target position or trading process in the low-level MDP. At timestamp $t$, if \(a_{lt} > P_t\), an ask order of predefined size will be placed. If \(a_{lt} < P_t\), a bid order of a predefined size will be placed. After that, \(P_{t+1}=a_{lt}\).

\noindent \textbf{Low-level Reward}, denoted as \(r_{lt} \in R_l\) at time $t$, is the net value difference between current time step and next one, which can be calculated as \(r_{lt}=(a_{lt} \times (p_{t+1}^{c}-p_t^{c}) - \delta \times |a_{lt}-P_t|) \times m\), where \(p_{t+1}^{c}\) and \(p_t^{c}\) are close prices, \(\delta\) is the transaction cost and \(m\) is the predefined holding size. 

\noindent \textbf{High-level State}, denoted as \(s_{ht} \in S_{h}\) at time $t$, consists of three parts: low-level features \(s_{ht}^1\), high-level context features \(s_{ht}^2\) and position state \(P_t\), where \(s_{ht}^1\) denotes low-level features, which is the combination of single-state features and low-level context features in low-level state, \(s_{ht}^2\) denotes high-level context features, which are the slope and volatility calculated over a backward window of length $h_c$ as shown in Section~\ref{sec:4.1}, \(P_t\) denotes the current position of the agent, which is the same as low-level MDP.

\noindent \textbf{High-level Action}, denoted as $a_{ht} \in A_{h}$ at time $t$, is the action of hyper-agent representing the target position of the trading process in the high-level MDP. Given a high-level state, the hyper-agent generates a softmax weight vector \(w=[w_1,...w_N]\), where $N$ is the number of sub-agents trained in low-level MDP. The final high-level action \(a_{ht} \in \{0,1\}\) is still the target position which is calculated as \(a_{ht}=\arg \max_{a'}(\sum_{i=1}^m w_i Q_i^{sub})\) where \(Q_i^{sub}\) denotes the output Q-value estimation of $i$-th sub-agent.

\noindent \textbf{High-level Reward}, denoted as \(r_{ht} \in R_h\), at time $t$ is the net value difference between the current time step and the next one, which is the same as low-level reward since our low-level and high-level MDPs operate under the same time scale. 

In our hierarchical MDP formulation, for every minute, sub-agents trained under different market dynamics provide their own decisions based on low-level states, and the hyper-agent executed in high-level MDP provides a final decision that takes all policies provided by sub-agents into consideration. Our goal is to train proper sub-agents and a hyper-agent to achieve the maximum accumulative profit. 

\section{MacroHFT}

\begin{figure*}[th]
\begin{center}
\includegraphics[width=\textwidth]{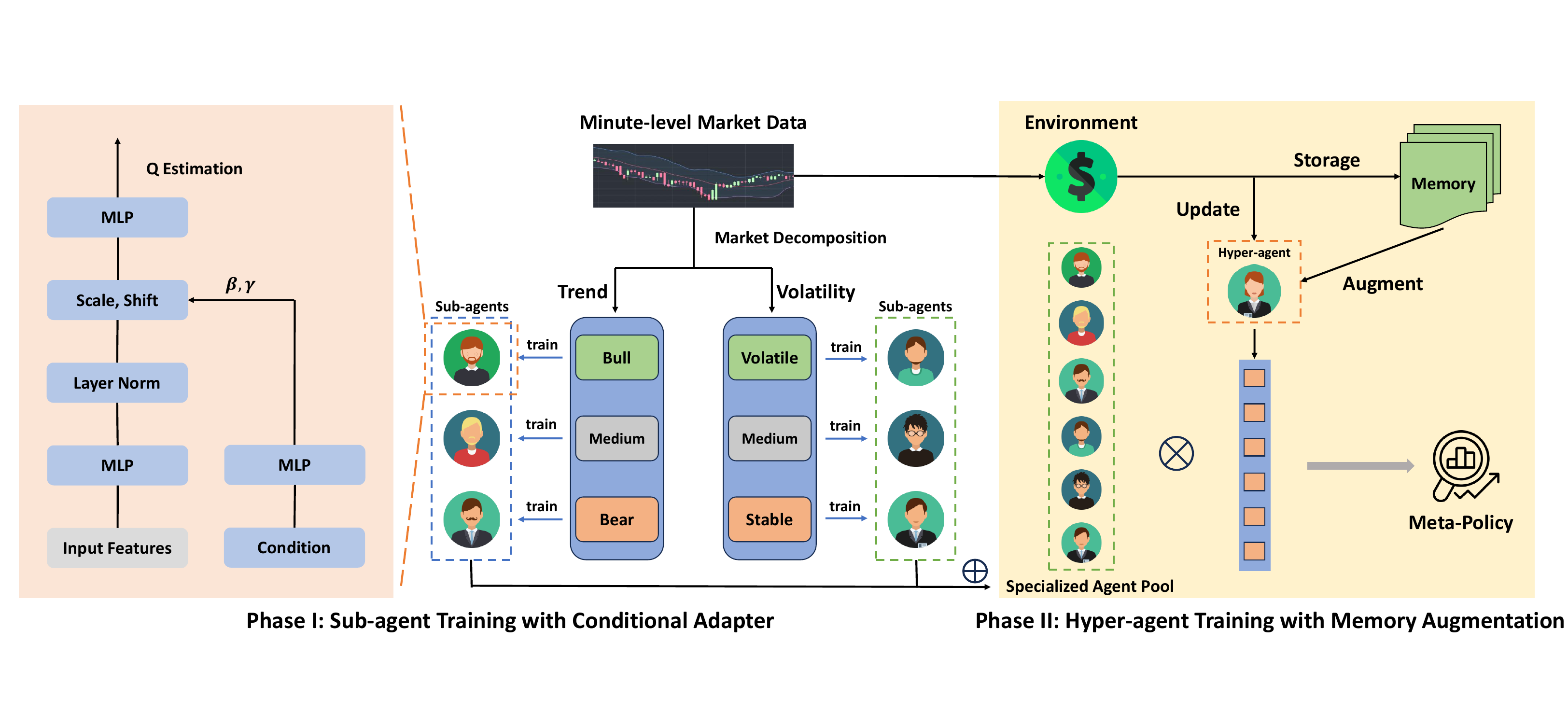}
\end{center}
\caption{The overview of MacroHFT. In phase I, we train multiple types of sub-agents with conditional adapters on the market data decomposed according to trend and volatility indicators. In phase II, we train a hyper-agent to mix decisions from all sub-agents, enhanced with a memory mechanism.}
\label{fig:overview}
\end{figure*}

In this section, we will introduce the detailed workflow of MacroHFT, which will be shown to be profitable in various non-stationary cryptocurrency markets.
As shown in Fig.~\ref{fig:overview}, MacroHFT mainly consists of two phases of RL training: 1) in phase one, MacroHFT will use conditioned RL method to train multiple sub-agents on low-level states tackling different market dynamics (markets of different trends and volatilities); 2) in phase two, MacroHFT will train a hyper-agent to provide a meta policy to fully 
exploit the potential of mixing diverse low-level policies based on recent market context.


\subsection{Market Decomposition}
\label{sec:4.1}

Because of data drifting caused by volatile cryptocurrency markets, it is usually impossible for a single RL agent to learn profitable trading policy from scratch over a long time period that contains various market conditions. We thus aim to train multiple sub-agents to execute policies diverse enough to tackle different market dynamics. 

Inspired by the market segmentation and labeling method introduced in \cite{qin2023earnhft}, we propose a market decomposition method based on the two most important market dynamic indicators: trend and volatility. 
In practice, given the market data that is a time series of OHLC prices and limit order book information, we will first segment the sequential data into chunks of fixed length $l_{chunk}$ for both the training set and validation set. 
Then we need to assign suitable trend and volatility labels for each chunk so that each sub-agent trained using data chunks belonging to the same market condition can handle a specific category of market dynamic. 
Specifically, 1) for trend labels, each data chunk will be first input into a low-pass filter for noise elimination. Then, a linear regression model is applied to the smoothed chunk, and the slope of the model is regarded as the indicator of market trend;
2) for volatility labels, the average volatility is calculated over each original chunk so that the fluctuations are maintained. 

In this case, each data block will be assigned the labels of two market dynamic indicators, including one trend label and one volatility label. Thus, all the data chunks can be divided into three subsets of equal length based on the quantiles of slope indicator and also three additional subsets based on the quantiles of volatility indicator, resulting in 6 training subsets containing data from bull (positive trend), medium (flat trend) and bear (negative trend) markets as well as volatile (high volatility), medium (flat volatility) and stable (low volatility) markets. After decomposing and labeling the training set, we further label the validation set using the quantile thresholds obtained from the training set so that we can perform fair evaluations of sub-agents over the markets they are expected to perform well on. By training an RL agent on each training subset and selecting the most profitable one based on the performance on the corresponding validation set, we are able to construct a total number of 6 trading sub-agents suitable for handling different market situations.  

\subsection{Low-Level Policy Optimization with Conditional Adaptation}
\label{sec:4.2}

Although previous works have stated the fact that value-based RL algorithms such as Deep Q-network have the ability to learn profitable policies for high-frequency cryptocurrency trading \cite{qin2023earnhft, zhu2022quantitative}, the trading agent's performance is largely influenced by overfitting issue \cite{zhang2023towards}. To be specific, the policy network might be too sensitive to some features or technical indicators while ignoring the recent market dynamics, which can lead to significant profit loss. Furthermore, the optimal policy of high-frequency trading largely depends on the current position of a trader due to the commission fee. Most existing trading algorithms try to include position information by simply concatenating it with state representations, but its effect on policy decision-making might be diminished because of its low dimension compared with state inputs. To tackle these challenges, we propose low-level policy optimization with conditional adaptation to train each sub-agent to learn adaptive low-level trading policies with conditional control. 

For sub-agent training, we use Double Deep Q-Network (DDQN) with dueling network architecture \cite{wang2016dueling} as our backbone and use context features $s_{lt}^2$ as well as current position $P_t$ as additional condition input to adapt output policy.
Given an input state tuple \( s_{lt}=(s_{lt}^1, s_{lt}^2, P_t) \) at timestamp $t$, where \(s_{lt}^1\), \(s_{lt}^2\), \(P_t\) denote single state features, context features and current position respectively, as defined in Section~\ref{Sec:3.2}, 
we employ two separate fully connected layers to extract semantic vectors of single and context features, and also a positional embedding layer to discrete position, which can be formulated as:
\begin{align}
    \begin{split}
     &h_s = \psi_1(s_{lt}^1), \\
     &c = \psi_3(P_t) + \psi_2(s_{lt}^2) \\
    \end{split}
\end{align}
where \(\psi_1\) and \(\psi_2\) denote the fully connected layers, and \(\psi_3\) denotes the positional embedding layer.
The obtained condition representation $c$ is constructed as the sum of the semantic vectors representing context and position information, and the single state is represented by its hidden embedding $h_s$.

Inspired by the Adaptive Layer Norm Block design in Diffusion Transformer \cite{peebles2023scalable}, we propose to adapt the single state representation \(h_s\) based on condition feature \(c\) so that the trained RL agent can generate suitable policies based on different market conditions and holding positions more efficiently. 
Thus, given the single state representation \(h_s \in R^{D}\), we first perform layer normalization across the whole hidden dimension, and then construct scale and shift vectors from condition vector \(c\) by linear transformation:
\begin{align}
&(\beta, \gamma) =\psi_c(c),
\end{align}
where the scale vector \( \beta \in R^D \), the shift vector \( \gamma \in R^D \), and \(\psi_c\) is a fully connected layer, and the adapted hidden state \(h \in R^D\) can be formed by 
\begin{align}
h=h_s \cdot \beta + \gamma, 
\end{align}
which serves as the input to the value and advantage network of DDQN to estimate Q values for each action as follows:
\begin{equation}
    Q^{sub}(h,a) = V(h) + (Adv(h,a) - \frac{1}{|A|}\sum_{a'\in A} Adv(h,a'))
\end{equation}
where \(V\) is the value network, \(Adv\) is the advantage network, \(A\) is the discrete action space. All network parameters are optimized by minimizing the one-step temporal-difference error as well as the Optimal Value Supervisor proposed in \cite{qin2023earnhft} which is the Kullback–Leibler (KL) divergence between the agent's Q estimation and optimal Q values (\(Q^*)\) calculated from dynamic programming of a given state. The loss function is constructed as follows:
\begin{align}
\begin{split}
    L = & ( r + \gamma Q_t^{sub}(h', \arg\max_{a'} Q^{sub}(h', a')) - Q^{sub}(h, a))^2 \\
        & + \alpha_{l} KL(Q^{sub}(h, \cdot)||Q^*)
\end{split}
\end{align}
where \(Q^{sub}\) is the policy network, \(Q_t^{sub}\) is the target network, \(Q^*\) is the optimal Q value, \(r\) is the reward, \(\gamma\) is the discount factor and \(\alpha\) is a coefficient controlling the importance of optimal Q supervisor. 

Overall, in order to generate diverse policies that are suitable for different market dynamics, 6 different sub-agents are trained using the above algorithm on 6 training subsets introduced in Section~\ref{sec:4.1}. The resulting low-level policies are further utilized to form the final trading policy by a hyper-agent, which will be introduced in the following section. 

\subsection{Meta-Policy Optimization with Memory Augmentation}

After learning diverse policies tackling different market conditions, we further train a hyper-agent that takes the decisions made by all sub-agents into consideration and outputs a high-level policy that can comfortably handle market dynamic changes and be consistently profitable. Specifically speaking, for a group of \(N\) optimized sub-agents with Q-value estimators denoted as \({Q^{sub}_1, Q^{sub}_2, ..., Q^{sub}_N}\) (N=6 in our setting), the hyper-agent outputs a softmax weight vector \(w=[w_1, w_2, ..., w_N]\) and aggregates decisions of sub-agents as a meta-policy function \(Q^{hyper}=\sum_{i=1}^N w_i Q^{sub}_i\), which fully leverages opinions from different sub-agents and prevents the meta trading policy from being highly one-sided. Moreover, to enhance the decision-making capability of the hyper-agent by correctly prioritizing sub-agents, a conditional adapter introduced in Section~\ref{sec:4.2} is also equipped, whose condition input is the slope and volatility indicators calculated over a backward window. 

However, standard RL optimization under the high-level MDP framework encounters several difficulties. Firstly, because of the rapid variation of cryptocurrency markets, the reward signals of similar states can vary largely, preventing the hyper-agent from learning a stable trading policy. Secondly, the performance of our meta-policy can be largely affected by extreme fluctuations that are rare and only last for a short time period, and the standard experience replay mechanism can hardly handle these situations. To handle these challenging issues, we propose an augmented memory that fully utilizes relevant experiences to learn a more robust and generalized meta-policy. 

Inspired by episodic memory used in many RL frameworks \cite{pritzel2017neural,lin2018episodic}, we construct a table-based memory module with limited storage capacity, denoted as \(M = (K, E, V)\), where \(K\) stores the key vectors that will be used for query, \(E\) stores the state and action pairs, and \(V\) stores the values. The usage of the memory module implies two operations: add and lookup. When a new episodic experience \(e=(s,a)\) and the resulting reward \(r\) is encountered, its corresponding key vector can be represented as its hidden state \(k=\psi_{enc}(s)\), where \(\psi_{enc}\) is the state encoder used in hyper-agent. The value of this experience can be calculated as the one-step Q estimation: \(v=r+\gamma max Q^{hyper}(s', \cdot)\), where \(Q^{hyper}\) is the action-value function of hyper-agent. Then, the obtained tuple \((k, (s,a), v)\) will be appended to the memory module. When the storage of the memory module reaches its maximum capacity, the experience tuple that is first added will be dropped, following a first-in-first-out mechanism. In this case, we can keep the memory with the most recent experiences that the hyper-agent encounters since they offer the most relevant knowledge to current decision-making. When conducting a lookup operation, we aim to retrieve the top-$m$ similar experiences stored in the memory and utilize the L2 distance between the vectors of the current hidden state and keys stored in the memory module to measure their similarity, formulated as:
\begin{equation}
    d(k, k_i)=||k-k_i||_2^2 + \epsilon, k_i \in K
\end{equation}
where \(\epsilon\) is a small constant. Then attention weight across the set of closest \(m\) experiences can be calculated as 
\begin{equation}
    w_i=\frac{d(k, k_i)\textbf{1}_{a=a_i}}{\sum_{i=1}^m d(k, k_i)\textbf{1}_{a=a_i}},
\end{equation}
and the aggregated value can be calculated as the weighted sum of values of these retrieved experiences with the same action taken at the current state:
\begin{equation}
    Q_M(s, a)=\sum\nolimits_{i=1}^{m}w_iv_i
\end{equation}
where \(v_i\) is the stored estimated value. 

While maintaining the standard RL target, we use this retrieved memory value \(Q_M\) as an additional target of the action-value estimation function in hyper-agent, and the loss function can be modified as follows:
\begin{align}
\begin{split}
    L = & (r + \gamma Q_t^{hyper}(s', \arg\max_{a'} Q^{hyper}(s', a')) - Q^{hyper}(s, a))^2 \\
        & + \alpha_{h} KL(Q^{hyper}(s, \cdot)||Q^*) + \beta (Q^{hyper}(s,a)-Q_M(s,a))^2
\end{split}
\end{align}

Through optimizing this objective, we aim to not only encourage the hyper-agent to enhance the consistency of its Q-value estimations across similar states but also allow the agent to quickly adapt its strategy in response to sudden market fluctuations. 

\begin{table*}[t]
\centering
\renewcommand{\arraystretch}{1.2}
  \resizebox{0.95\textwidth}{!}{
\begin{tabular}{|c|c|c|ccc|cc|c|c|c|c|ccc|cc|c|}
\hline
\multicolumn{2}{ |c | }{} & \multicolumn{1}{ c | }{Profit} & \multicolumn{3}{ c | }{Risk-Adjusted Profit} & \multicolumn{2}{ c| }{Risk Metrics}  & \multicolumn{1}{ c | }{Trading}  &\multicolumn{2}{ c | }{} & \multicolumn{1}{ c | }{Profit} & \multicolumn{3}{ c | }{Risk-Adjusted Profit} & \multicolumn{2}{ c| }{Risk Metrics}  & \multicolumn{1}{ c | }{Trading}  
\\ 
\cline{1-8} \cline{10-17}
Market   & Model & TR(\%)$\uparrow$ & ASR$\uparrow$ & ACR$\uparrow$ & ASoR$\uparrow$ & AVOL(\%)$\downarrow$ & MDD(\%)$\downarrow$ & Number &Market   & Model & TR(\%)$\uparrow$ & ASR$\uparrow$ & ACR$\uparrow$ & ASoR$\uparrow$ & AVOL(\%)$\downarrow$ & MDD(\%)$\downarrow$ & Number \\
\hline
{\multirow{7}{*}{\rotatebox[origin=c]{0}{BTC}}} & CLSTM-PPO & -10.67& -0.92   & -1.38 & -0.85 &   32.96  & 22.01 & 20 & {\multirow{7}{*}{\rotatebox[origin=c]{0}{DOT}}} & CLSTM-PPO & \textcolor[RGB]{0,128,255}{-2.41} & {-2.86}    & {-2.27} & -0.10 &   \textcolor[RGB]{255,0,185}{2.03} & \textcolor[RGB]{255,0,185}{2.56} & 59 \\
&  PPO  & {-9.15} & \textcolor[RGB]{0,153,76}{-0.75}   & \textcolor[RGB]{0,153,76}{-1.15} & \textcolor[RGB]{0,128,255}{-0.69} &   33.29  & {21.66}& 1& & PPO  &  -5.42  & -3.00 & -2.24 & {-0.09} & {4.41} &  {5.91} & 55\\
\cline{2-9} \cline{11-18}

& CDQNRP & \textcolor[RGB]{0,153,76}{-1.51} & {-3.74} &{-2.45}& \textcolor[RGB]{0,153,76}{-0.28} & \textcolor[RGB]{255,0,185}{1.29} &  \textcolor[RGB]{255,0,185}{1.97} & 75& &
CDQNRP &{-3.20} & -1.87 & -1.86& -0.10 & \textcolor[RGB]{0,128,255}{4.11}  & \textcolor[RGB]{0,128,255}{4.14}  & 139\\ 
& DQN & -10.41 & \textcolor[RGB]{0,128,255}{-0.90} & \textcolor[RGB]{0,128,255}{-1.34} & -0.83 & {32.87} &   {21.97} & 58 && DQN & {-4.99} & {-5.18} & {-2.25} & {-0.22} & \textcolor[RGB]{0,153,76}{2.35}  & {5.42} & 106\\ 
& DDQN & \textcolor[RGB]{0,128,255}{-9.14} & {-11.52} & {-3.22} & -0.96 & \textcolor[RGB]{0,153,76}{2.77} &   \textcolor[RGB]{0,128,255}{9.91} & 282 && DDQN & {-3.75} & {-2.19} & {-2.23} & {-0.08} & {4.13}  & \textcolor[RGB]{0,153,76}{4.05} & 111\\ 
\cline{2-9} \cline{11-18}

& MACD &   -18.99 & -3.07 & -2.86 & -2.06 & {21.03} &   22.57 &234 && MACD &  -20.29 & -1.52 &  -1.65 & -0.91 & 32.19  & 29.74 & 277\\ 
& IV &{-9.24}  & {-1.57} & {-1.99} & {-0.93} & {18.50}  & {14.62} & 120&& IV & \textcolor[RGB]{0,153,76}{10.58} & \textcolor[RGB]{255,0,185}{1.01} & \textcolor[RGB]{0,153,76}{1.53} & \textcolor[RGB]{0,153,76}{0.58} & 27.70 &  18.26 & 88\\ 
\cline{2-9}\cline{11-18}
& EarnHFT & {-11.16} & {-0.96} & {-1.45} & {-0.89}& {33.41}  &{22.08} &23 && EarnHFT & {-2.67} & \textcolor[RGB]{0,128,255}{-0.98} & \textcolor[RGB]{0,128,255}{-1.09} &\textcolor[RGB]{0,128,255}{-0.01}&6.40  &5.80 &17\\
\cline{2-9}\cline{11-18}
& MacroHFT & \textcolor[RGB]{255,0,185}{3.03} & \textcolor[RGB]{255,0,185}{0.61} & \textcolor[RGB]{255,0,185}{2.06} & \textcolor[RGB]{255,0,185}{0.34}& \textcolor[RGB]{0,128,255}{18.19}  &\textcolor[RGB]{0,153,76}{5.41} &19  && MacroHFT & \textcolor[RGB]{255,0,185}{13.79} & \textcolor[RGB]{0,153,76}{0.97} & \textcolor[RGB]{255,0,185}{2.45} &\textcolor[RGB]{255,0,185}{0.68}&40.31  &15.89 &38\\
\hline

{\multirow{7}{*}{\rotatebox[origin=c]{0}{ETH}}} &  CLSTM-PPO & -17.87 & {-1.20}   & -1.23 & -1.14 &   34.23 & 33.56 & 407 & {\multirow{7}{*}{\rotatebox[origin=c]{0}{LTC}}} & CLSTM-PPO & -24.96 &-0.70   & -0.93 & -0.61 &   66.39  & 50.00 & 1\\
&  PPO  & -2.12 & 0.05   & {0.08} &  0.05 &   37.44  & 24.76 & 1 && PPO  & -24.96 &-0.70   & -0.93 & -0.61 &   66.39  & 50.00 & 1\\
\cline{2-9}\cline{11-18}
 & CDQNRP & {-2.30} & 0.04 & \textcolor[RGB]{0,128,255}{0.6} & \textcolor[RGB]{0,128,255}{0.4} & {37.43}  & {24.75} &3 & & CDQNRP & -1.72 & -1.19 & -2.37 & -0.05 & \textcolor[RGB]{255,0,185}{3.45} &  \textcolor[RGB]{255,0,185}{1.73} &63\\ 
& DQN & -4.14 & -0.09 & -0.13 & -0.08 & {36.92}  & {25.59} &7 && DQN & -3.26 & -1.00 & -1.35 & -0.01 & \textcolor[RGB]{0,153,76}{7.62} & \textcolor[RGB]{0,128,255}{5.65} &14 \\
& DDQN & -8.72 & -0.43 & -0.54 & -0.41 & {35.71}  & {28.52} &111 && DDQN & -1.74 & -0.34 & -0.69 & -0.01 & {10.66} & \textcolor[RGB]{0,153,76}{5.22} &130 \\
\cline{2-9}\cline{11-18}

& MACD & {-7.96} & {-0.72} & {-0.75} & {-0.49} &  \textcolor[RGB]{0,128,255}{23.63}  & 22.86 &286& & MACD &  -13.16 & -0.72 & -1.03 &  -0.46 &   {37.11} &    26.00 & 272\\ 
& IV & \textcolor[RGB]{0,128,255}{0.56} & \textcolor[RGB]{0,128,255}{0.17} & 0.32 &  0.09 & \textcolor[RGB]{255,0,185}{19.48}  & \textcolor[RGB]{0,153,76}{9.98} &80 && IV & \textcolor[RGB]{0,153,76}{7.75} & \textcolor[RGB]{0,153,76}{0.76} & \textcolor[RGB]{0,153,76}{1.13} &  \textcolor[RGB]{0,153,76}{0.40} & 28.83  & 19.47 &92\\ \cline{2-9}\cline{11-18}
 & EarnHFT & \textcolor[RGB]{0,153,76}{18.02} &\textcolor[RGB]{0,153,76}{1.53}  &\textcolor[RGB]{0,153,76}{3.59}  &\textcolor[RGB]{0,153,76}{1.23}& {28.60}  &\textcolor[RGB]{0,128,255}{12.21} &270& & EarnHFT & \textcolor[RGB]{0,128,255}{0.54} &\textcolor[RGB]{0,128,255}{0.16}  & \textcolor[RGB]{0,128,255}{0.30}  &\textcolor[RGB]{0,128,255}{0.01}&  \textcolor[RGB]{0,128,255}{17.80}  & {9.63} &16  \\
 \cline{2-9}\cline{11-18}
  & MacroHFT & \textcolor[RGB]{255,0,185}{39.28} &\textcolor[RGB]{255,0,185}{3.89}  &\textcolor[RGB]{255,0,185}{8.41}  &\textcolor[RGB]{255,0,185}{2.49}& \textcolor[RGB]{0,153,76}{20.93}  &\textcolor[RGB]{255,0,185}{9.67} &20 & & MacroHFT & \textcolor[RGB]{255,0,185}{18.16} &\textcolor[RGB]{255,0,185}{1.50}  & \textcolor[RGB]{255,0,185}{3.11}  &\textcolor[RGB]{255,0,185}{0.66}&  29.59  & {14.24} &138 \\
\hline
\end{tabular}
}
\caption{Performance comparison on 4 crypto markets with 8 baselines including 2 policy-based RL, 3 value-based RL, 2 rule-based and 1 hierarchical RL algorithms. Results in pink, green, and blue show the best, second-best, and third-best results.}
\label{tab:performance_tex}
\end{table*}

\section{Experiments}
\subsection{Datasets}

To comprehensively evaluate the effectiveness of our proposed MacroHFT, experiments are conducted on four cryptocurrency markets, where the training, validation and test subset splitting is shown in Table~\ref{tab:1}. We first decompose and label the train and validation set based on market trend and volatility using the method described in Section~\ref{sec:4.1}. Then, we train a separate sub-agent on data chunks with different labels in the training set and conduct model selection based on the sub-agent's mean return rate on the validation set. We further train the hyper-agent over the whole training set and pick the best one according to its total return rate on the whole validation set. 
\begin{table}[ht]
\centering
\footnotesize
\begin{tabular}{llll}
\toprule
\multicolumn{1}{c}{\textbf{Dataset}} & \multicolumn{1}{c}{\textbf{Train}} & \multicolumn{1}{c}{\textbf{Validation}} & \multicolumn{1}{c}{\textbf{Test}} \\
\midrule
BTC/USDT & 22/03/05 - 23/02/22 & 23/03/18 - 23/06/15 & 23/06/22 - 23/10/15 \\
ETH/USDT & 22/02/01 - 23/01/31 & 23/02/01 - 23/05/31 & 23/06/01 - 23/10/31 \\
DOT/USDT & 22/02/01 - 23/01/31 & 23/02/01 - 23/05/31 & 23/06/01 - 23/10/31 \\
LTC/USDT & 22/02/01 - 23/01/31 & 23/02/01 - 23/05/31 & 23/06/01 - 23/10/31 \\
\bottomrule
\\
\end{tabular}
\caption{Datasets and data splits for four cryptocurrency markets}
\label{tab:1}
\end{table}

\subsection{Evaluation Metrics}

We evaluate our proposed method on 6 different financial metrics including one profit criterion, two risk criteria, and three risk-adjusted profit criteria listed below.
\begin{itemize}
    \item \textbf{Total Return (TR)} is the overall return rate of the entire trading period, which is defined as \(TR=\frac{V_T-V_1}{V_1}\), where \(V_T\) is the final net value and \(V_1\) is the initial net value.
    \item \textbf{Annual Volatility (AVOL)} is the variation in an investment’s return over one year measured as \(\sigma[r] \times \sqrt{m}\), where \(r=[r_1, r_2, ..., r_T]\) is the return vector of every minute, \(\sigma[\cdot]\) is the standard deviation, \(m=525600\) is the number of minutes in a year.
    \item \textbf{Maximum Drawdown (MDD)} measures the largest loss from any peak to show the worst case. 
    \item \textbf{Annual Sharpe Ratio (ASR)} measures the amount of extra return that a trader receives per unit of increased risk, calculated as \(ASR=E[r] / \sigma[r] \times \sqrt{m}\) where \(E[\cdot]\) is the expectation. 
    \item \textbf{Annual Calmar Ratio (ACR)} measures the risk-adjusted return calculated as \(ACR=\frac{E[r]}{MDD} \times m\).
    \item \textbf{Annual Sortino Ratio (ASoR)} applies the downside deviation (DD) as the risk measure, which is defined as \(SoR=\frac{E[r]}{DD} \times \sqrt{m}\), where DD is the standard deviation of the negative return rates. 

\end{itemize}
{
\setlength{\abovecaptionskip}{2pt}
\setlength{\belowcaptionskip}{0pt}
\begin{figure}[t]
\centering
\includegraphics[width=0.46\textwidth]{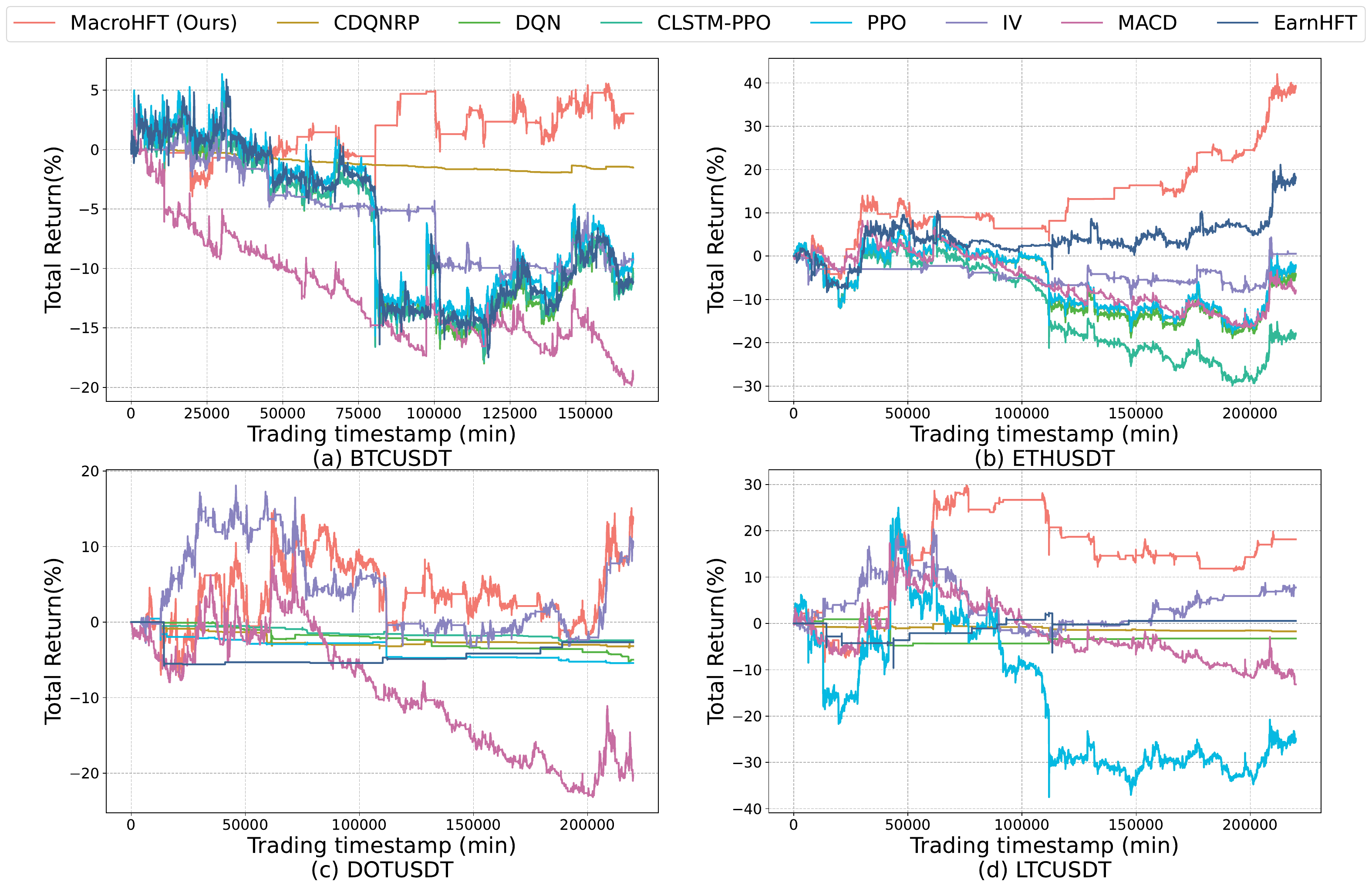}
\caption{Performance of MacroHFT and other baselines}
\label{fig:baseline_curve}
\end{figure}
}
\begin{figure}[tbp]
\centering
\includegraphics[width=0.46\textwidth]{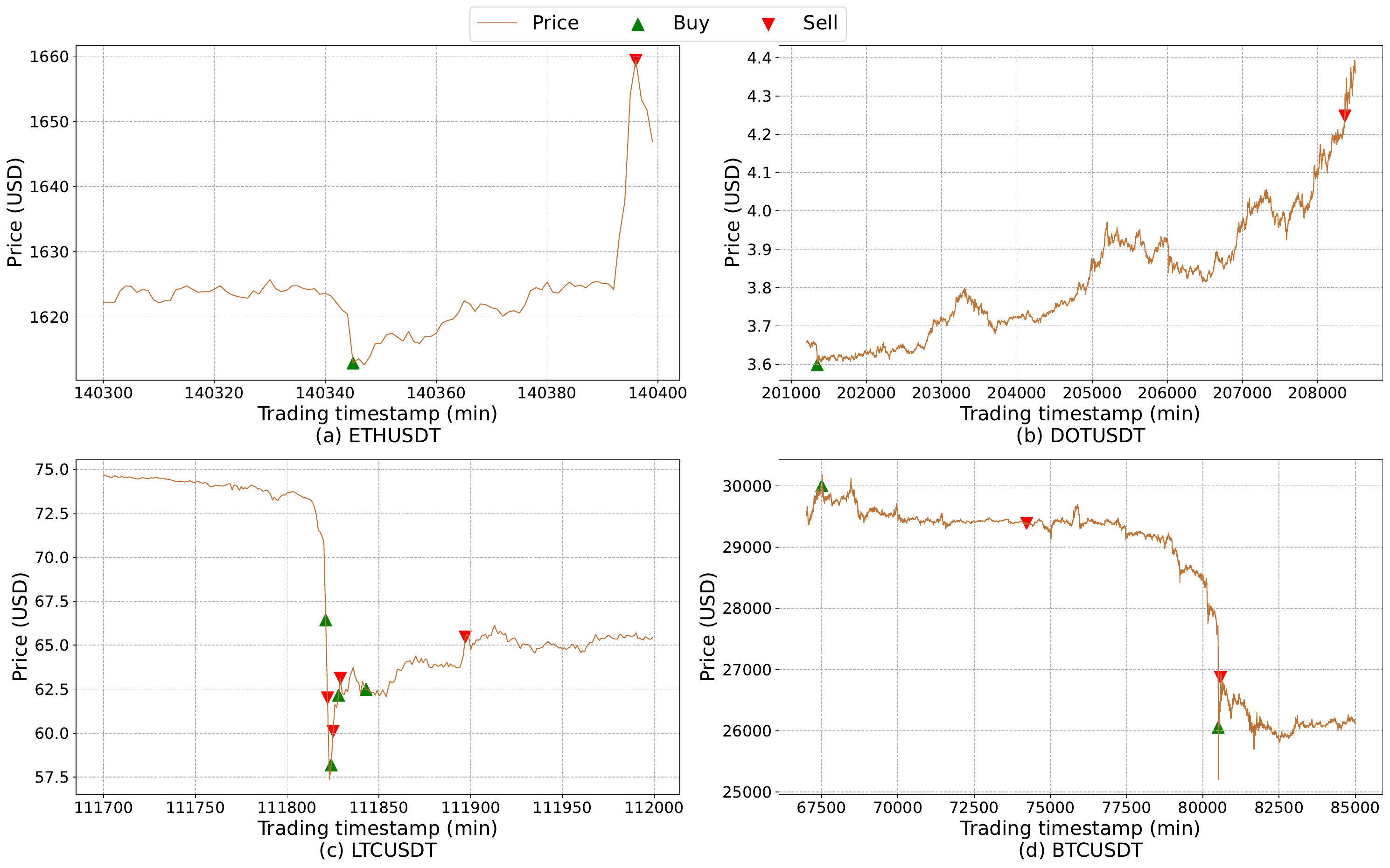}
\caption{Trading examples of different cryptocurrencies}
\label{fig:trading_example}
\end{figure}

\subsection{Baselines}
To provide a comprehensive comparison of our proposed method, we select 8 baselines including 6 SOTA RL algorithms and 2 widely-used rule-based trading strategies.
\begin{itemize}
    \item \textbf{DQN} \cite{mnih2015human} is a value-based RL algorithm applying experience replay and multi-layer perceptrons to Q-learning. 
    \item \textbf{DDQN} \cite{wang2016dueling} is a modification of DQN which uses a separate target network for selecting and evaluating actions to reduce the overestimation bias in action value estimates. 
    \item \textbf{PPO} \cite{schulman2017proximal} is a policy-based RL algorithm that balances the trade-off between exploration and exploitation by clipping the policy update function, which enhances training stability and efficiency.
    \item \textbf{CDQNRP} \cite{zhu2022quantitative} is a modification of DQN which uses a random perturbed target frequency to enhance the stability during training.
    \item \textbf{CLSTM-PPO} \cite{zou2024novel} is a modification of PPO which uses LSTM to enhance state representation. 
    \item \textbf{EarnHFT} \cite{qin2023earnhft} is a hierarchical RL framework that trains low-level agents on different market trends and a router to select suitable agents based on macro market information. 
    \item \textbf{IV} \cite{chordia2002order} is a micro-market indicator reflecting short-term market direction which is widely used in HFT.
    \item \textbf{MACD} \cite{krug2022enforcing} is a modification of the traditional moving average method considering both direction and changing speed of the current price.
\end{itemize}

\begin{figure}[t]
\centering
\includegraphics[width=0.45\textwidth]{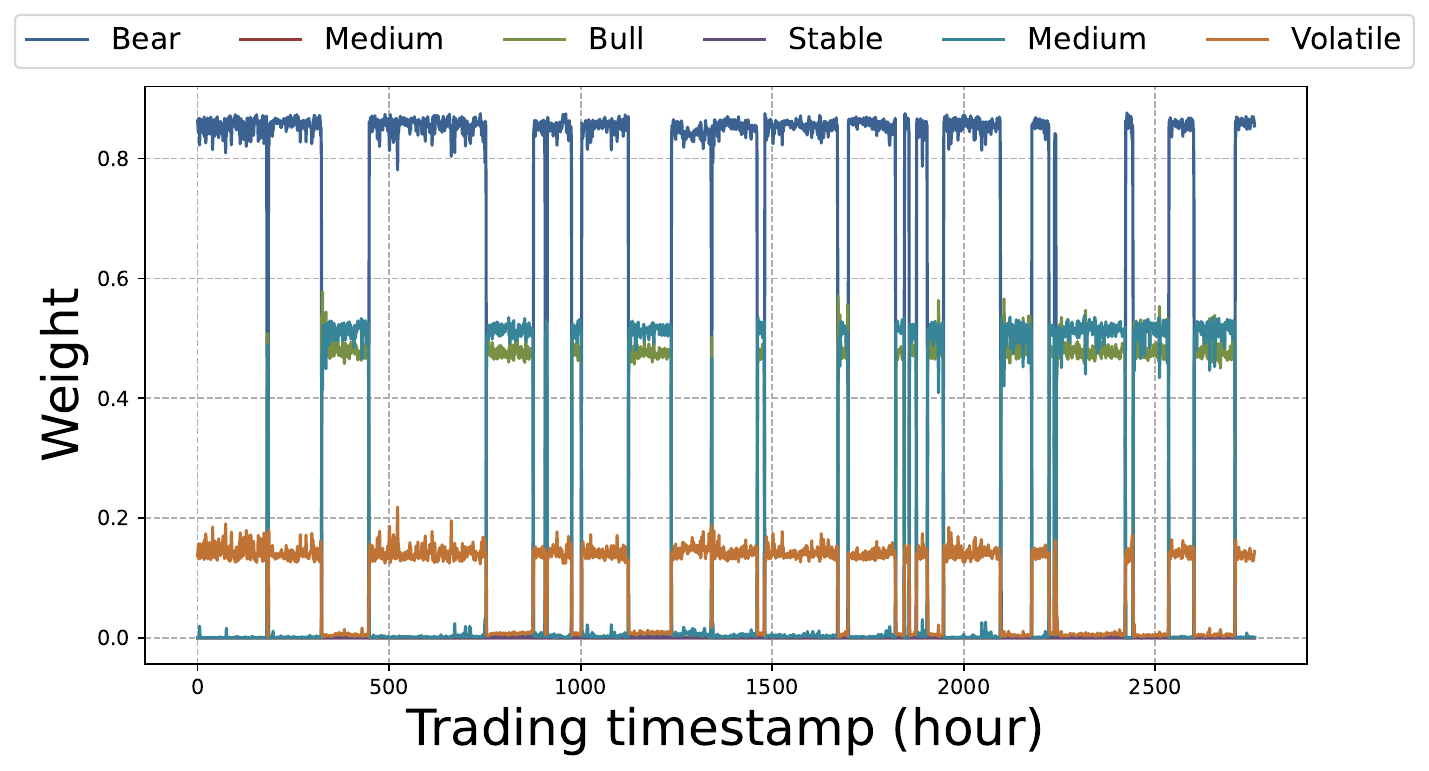}
\caption{Weight of sub-agents assigned by hyper-agent in BTCUSDT}
\label{fig:weight_btc}
\end{figure}

\subsection{Experiment Setup}

We conduct all experiments on 4 4090 GPUs. For the trading setting, the commission fee rate is 0.02\% for all four cryptocurrencies following the policy of Binance. For sub-agent training, the embedding dimension is 64 and the policy network's dimension is 128. The decomposed data chunk length $l_{chunk}$ is explored over \(\{360, 4320\}\) \footnote{6 hours and 3 days}. For each dataset, we conduct both training phases and determine \(l_{chunk}\) based on the overall return rate of meta-policy over the validation sets. For BTCUSDT, \(l_{chunk}\) is set as 360. For the other three datasets, \(l_{chunk}\) is set as 4320. All the sub-agents are trained for 15 epochs and selected based on the average return rate on the corresponding validation subsets with the same market label. The coefficient \(\alpha_l\) of each sub-agent is tuned separately over \(\{0,1,4\}\) and selected based on the mean return rate of the validation subset with the same label of the agent. For hyper-agent training, the embedding dimension is 32 and the policy network's dimension is 128. The hyper-agent is trained for 15 epochs and selected based on the return rate over the whole validation set. All the network parameters are optimized by Adam optimizers with a learning rate of 1e-4. The coefficient \(\alpha_h\) is set to be 0.5, and \(\beta\) is tuned over \(\{1,5\}\) and selected based on the overall return rate of meta-policy over the validation set. For DOTUSDT, \(\beta\) is set as $1$. For the other three datasets, \(\beta\) is set as $5$.

\subsection{Results and Analysis}

The performance of MacroHFT and other baseline methods on 4 cryptocurrencies are shown in Table~\ref{tab:performance_tex} and Figure~\ref{fig:baseline_curve}. It can be observed that our method achieves the highest profit and the highest risk-adjusted profit in all 4 cryptocurrency markets for most of the evaluation metrics. Furthermore, although chasing for larger potential profit can lead to higher risk, MacroHFT still performs competently in risk management compared with baseline methods. For baseline comparisons, value-based methods (CDQNRP, DQN) demonstrate consistent performance across a majority of datasets; however, they fall short in generating profit. Policy-based methods (PPO, CLSTM-PPO) exhibit high sensitivity during the training process and can easily converge to simplistic policies (\emph{e.g.} buy-and-hold), resulting in poor performance, especially in bear markets. Certain rule-based methods (\emph{e.g.,} IV) can yield profit on most of the datasets. However, their success heavily relies on the precise tuning of the take-profit and stop-loss thresholds, which necessitates the input of human expertise. Nevertheless, there are also rule-based trading strategies (\emph{e.g.,} MACD) that perform poorly across numerous datasets, leading to significant losses. The hierarchical RL method (EarnHFT) achieves good performance on both profit-making and risk management over two datasets but fails to make profits on the other datasets. 

To look into more detailed trading strategies of MacroHFT, we visualize some actual trading signal examples in different cryptocurrency markets, which are shown in Figure~\ref{fig:trading_example}. From the trading example in the ETH market (Figure~\ref{fig:trading_example}(a)), it can be observed that by executing a potential "breakout" strategy, MacroHFT successfully seizes the fleeting opportunity of making profits. This indicates that our MacroHFT is able to respond rapidly to momentary market fluctuations and make profits in short intervals, which is the common goal of high-frequency trading. From the trading example in the DOT market (Figure~\ref{fig:trading_example}(b)), it is apparent that MacroHFT executes a trend-following strategy over a long period of bull markets and exits its position after gaining a substantial profit. It is evident that with the help of conditional adaptation, our MacroHFT also shows great capacity of grabbing significant market trends and achieving better long-term returns. From the trading example in the LTC market (Figure~\ref{fig:trading_example}(c)), it can be observed that MacroHFT executes a stop-loss action when encountering a collapse and makes profits when the market rebounds. In the trading example in the BTC market (Figure~\ref{fig:trading_example}(d)), MacroHFT still manages to seize the opportunity of small advances even in a bear market, indicating the robustness of our method under adverse conditions. Furthermore, an example of the hyper-agent's weight assignment of different sub-agents in the BTC market is also displayed. From the curves representing the average weight changes of sub-agents in a 60-minute interval (Figure~\ref{fig:weight_btc}), we can notice that MacroHFT successfully generates consistently profitable trading strategies by mixing decisions reasonably from different sub-agents based on various market conditions, while it remains the ability to adjust quickly to sudden market changes.  

\begin{table}[!b]
\centering
\footnotesize
\begin{tabular}{lcccccc}
\toprule
& \multicolumn{3}{c}{BTCUSDT} & \multicolumn{3}{c}{ETHUSDT} \\
\cmidrule(r){2-4} \cmidrule(l){5-7}
Model & TR(\%)$\uparrow$ & ASR$\uparrow$ & MDD(\%)$\downarrow$ & TR(\%)$\uparrow$ & ASR$\uparrow$ & MDD(\%)$\downarrow$ \\
\midrule
w/o-CA & 1.69 & 0.36 & 7.24 & 14.27 & 2.42 & \underline{7.57} \\
w/o-MEM & 2.03 & 0.49 & 6.49 & 12.73 & 1.26 & 20.87 \\
MacroHFT & \underline{3.03} & \underline{0.61} & \underline{5.41} & \underline{39.28} & \underline{3.89} & {9.67} \\
\bottomrule
\end{tabular}

\begin{tabular}{lcccccc}
\toprule
& \multicolumn{3}{c}{DOTUSDT} & \multicolumn{3}{c}{LTCUSDT} \\
\cmidrule(r){2-4} \cmidrule(l){5-7}
Model & TR(\%)$\uparrow$ & ASR$\uparrow$ & MDD(\%)$\downarrow$ & TR(\%)$\uparrow$ & ASR$\uparrow$ & MDD(\%)$\downarrow$ \\
\midrule
w/o-CA & -16.79 & -1.18 & 31.66 & -6.71 & -0.36 & 18.83 \\
w/o-MEM & 2.41 & 0.34 & 27.23 & -8.66 & -0.58 & 22.03 \\
MacroHFT & \underline{13.79} & \underline{0.97} & \underline{15.89} & \underline{18.16} & \underline{1.50} & \underline{14.24} \\
\bottomrule
\end{tabular}
\caption{Performance comparison of models across four datasets. Underlined results represent the best performance}
\label{tab:ablation}
\end{table}

{
\setlength{\abovecaptionskip}{2pt}
\setlength{\belowcaptionskip}{0pt}
\begin{figure}[t]
\centering
\includegraphics[width=0.47\textwidth]{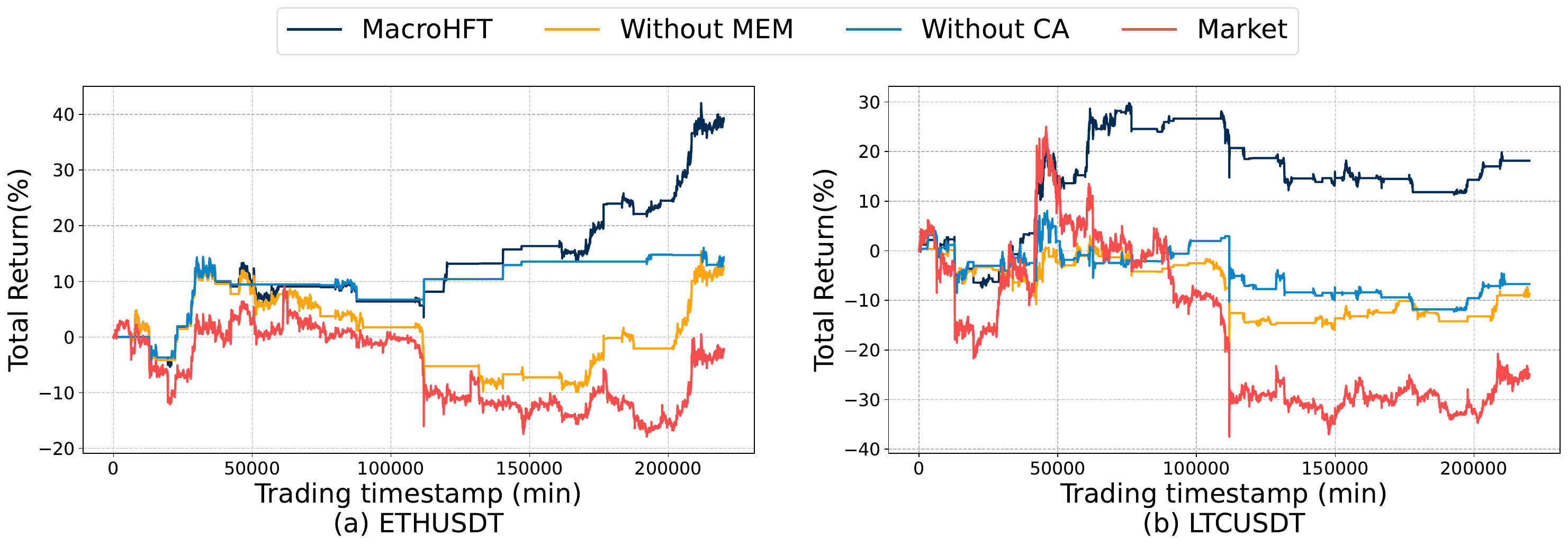}

\caption{Performance of original MacroHFT and two variations without conditional adapter and memory}
\label{fig:ablation_curve}
\end{figure}
}

\subsection{Ablation Study}
\noindent To investigate the effectiveness of our proposed conditional adapter (CA) and memory (MEM), ablation experiments are conducted by removing respective modules and the results are displayed in Table~\ref{tab:ablation}. It can be observed that the original MacroHFT with both conditional adapter and memory achieves the highest profit, the highest risk-adjusted profit and the lowest investment risk except for the MDD criterion of the ETH market. This indicates that both conditional adapter and memory play important roles in generating more profitable trading strategies and controlling investment risks. For harsh trading environments such as DOTUSDT and LTCUSDT markets, where market values decrease by 14.85 \% and 24.94\% respectively, the removal of these two modules can cause significant deficit. 

Furthermore, We can gain a more intuitive understanding of the influence of conditional adapter and memory modules on hyper-agent's trading behavior from Figure~\ref{fig:ablation_curve}, which is the return rate curves of different ablation models in ETH and LTC markets. Referring to Figure~\ref{fig:ablation_curve}, it can be observed that MacroHFT without memory cannot reply timely to the sudden fall in the ETH market which leads to a huge loss. At the same time, MacroHFT without conditional adapter fails to adjust its trading strategy when the market trend switches from flat or bear to bull, missing the chance to make more profit. Meanwhile, our proposed MacroHFT with both conditional adapter and memory achieves strong performance under different types of markets because of its ability to adjust policy based on market context and react promptly to abrupt fluctuations.

\section{Conclusion}
In this paper, we propose MacroHFT, a novel memory-augmented context-aware RL method for HFT. Firstly, we train different types of sub-agents with market data decomposed according to the market trend and volatility for better specialization capacity. Agents are also equipped with conditional adapters to adjust their trading policy according to market context, preventing them from overfitting. Then, we train a hyper-agent to blend decisions from different sub-agents for less biased trading strategies. A memory mechanism is also introduced to enhance the hyper-agent's decision-making ability when facing precipitous fluctuations in cryptocurrency markets. Comprehensive experiments across various cryptocurrency markets demonstrate that MacroHFT significantly surpasses multiple state-of-the-art trading methods in profit-making while maintaining competitive risk managing ability, and achieves superior performance on minute-level trading tasks. 

\section{Acknowledgments}
This project is supported by the National Research Foundation, Singapore under its Industry Alignment Fund – Pre-positioning (IAF-PP) Funding Initiative. Any opinions, findings and conclusions or recommendations expressed in this material are those of the author(s) and do not reflect the views of National Research Foundation, Singapore.


\bibliographystyle{ACM-Reference-Format}
\balance
\bibliography{sample-base}

\clearpage
\appendix

\section{Details of Technical Indicators}
\label{appendix}
In this section, we elaborate on the details of technical indicators used in MacroHFT mentioned in Section~\ref{Sec:3.1}. The definitions and calculation formulas of technical indicators are shown in Table~\ref{tab:indicators}.

\begin{table*}[th]
\begin{tabular}{|c|c|}
\hline
\textbf{Indicator} & \textbf{Calculation Formula} \\
\hline
max\_oc & $y_{\text{max\_oc}} = \max(p_t^o, p_t^c)$ \\
min\_oc & $y_{\text{min\_oc}} = \min(p_t^o, p_t^c)$ \\
kmid & $y_{\text{kmid}} = (p_t^c - p_t^o)$ \\
kmid2 & $y_{\text{kmid2}} = (p_t^c - p_t^o) / (p_t^h - p_t^l)$ \\
klen & $y_{\text{klen}} = (p_t^h - p_t^l)$ \\
kup & $y_{\text{kup}} = (p_t^h - y_{\text{max\_oc}})$ \\
kup2 & $y_{\text{kup2}} = (p_t^h - y_{\text{max\_oc}}) / (p_t^h - p_t^l)$ \\
klow & $y_{\text{klow}} = (y_{\text{min\_oc}} - p_t^l) / (p_t^h - p_t^l)$ \\
klow2 & $y_{\text{klow2}} = (y_{\text{min\_oc}} - p_t^l) / (p_t^h - p_t^l)$ \\
ksft & $y_{\text{ksft}} = (2 \times p_t^c - p_t^h - p_t^l)$ \\
ksft2 & $y_{\text{ksft2}} = (2 \times p_t^c - p_t^h - p_t^l) / (p_t^h - p_t^l)$ \\
volume & $y_\text{volume} = \sum_{i=1}^5 (q_{t}^{b_i}+q_{t}^{a_i})$ \\
\hline
\multirow{2}{*}{{bid$_\text{i}$\_size\_n}} & $i \in I = [1,2,3,4,5]$ \\
& $\text{bid$_\text{i}$\_size\_n} = q_{t}^{b_i} / y_\text{volume}$ \\
\hline
\multirow{2}{*}{{ask$_\text{i}$\_size\_n}} & $i \in I = [1,2,3,4,5]$ \\
& $\text{ask$_\text{i}$\_size\_n} = q_{t}^{a_i} / y_\text{volume}$ \\
\hline
wap1 & $y_{\text{wap1}} = (q_{t}^{a_1}*p_{t}^{b_1}+q_{t}^{b_1}*p_{t}^{a_1})/(q_{t}^{a_1}+q_{t}^{b_1})$ \\
wap2 & $y_{\text{wap2}} = (q_{t}^{a_2}*p_{t}^{b_2}+q_{t}^{b_2}*p_{t}^{a_2})/(q_{t}^{a_2}+q_{t}^{b_2})$ \\
wap\_balance & $y_{\text{wap\_balance}} = |y_{\text{wap1}}-y_{\text{wap2}}|$ \\
buy\_spread & $y_{\text{buy\_spread}} = |p_{t}^{b_1}-p_{t}^{b_5}|$ \\
sell\_spread & $y_{\text{sell\_spread}} = |p_{t}^{a_1}-p_{t}^{a_5}|$ \\
buy\_volume & $y_{\text{buy\_volume}} = \sum_{i=1}^5 q_{t}^{b_i}$ \\
sell\_volume & $y_{\text{sell\_volume}} = \sum_{i=1}^5 q_{t}^{a_i}$ \\
volume\_imbalance & $y_{\text{volume\_imbalance}} = {(y_{\text{buy\_volume}}-y_{\text{sell\_volume}})}/{(y_{\text{buy\_volume}}+y_{\text{sell\_volume}})}$\\
price\_spread & $y_{\text{price\_spread}} = {2*(p_{t}^{a_1}-p_{t}^{b_1})}/{(p_{t}^{a_1}+p_{t}^{b_1})}$ \\
sell\_vwap & $y_{\text{sell\_vwap}} = \sum_{i=1}^5 ask_{i}\_size\_n * p_{t}^{a_i}$ \\
buy\_vwap & $y_{\text{buy\_vwap}} = \sum_{i=1}^5 bid_{i}\_size\_n * p_{t}^{b_i}$ \\
\hline
\multirow{2}{*}{{log\_return\_bid$_\text{i}$\_price}} & $i \in I = [1,2]$ \\
& $\text{log\_return\_bid$_\text{i}$\_price} = log(p_{t}^{b_i}/p_{t-1}^{b_i})$ \\
\hline
\multirow{2}{*}{{log\_return\_ask$_\text{i}$\_price}} & $i \in I = [1,2]$ \\
& $\text{log\_return\_ask$_\text{i}$\_price} = log(p_{t}^{a_i}/p_{t-1}^{a_i})$ \\
\hline
log\_return\_wap1 & $y_{\text{log\_return\_wap1}} = log(y_{wap1}^{t}/y_{wap1}^{t-1})$ \\
log\_return\_wap2 & $y_{\text{log\_return\_wap2}} = log(y_{wap2}^{t}/y_{wap2}^{t-1})$ \\
\hline
\multirow{2}{*}{{trend\_features}} &   
$y \in Y = [p_{t}^{a_i}, p_{t}^{b_i}, y_{\text{buy\_spread}}, y_{\text{sell\_spread}}, y_{\text{wap1}}, y_{\text{wap2}}, y_{\text{sell\_vwap}}, y_{\text{buy\_vwap}}, y_{\text{volume}}]$ \\  
& $y_{\text{trend}} = {y - \text{RollingMean}(y, 60)}/{\text{RollingStd}(y, 60)}$ \\
\hline 
\end{tabular}
\caption{Calculation Formulas for Indicators}
\label{tab:indicators}
\end{table*}

\end{document}